# A novel robot calibration method with plane constraint based on dial indicator

Tinghui Chen, Shuai Li, *Senior Member, IEEE* and Hao Wu

*Abstract*—In pace with the electronic technology development and the production technology improvement, industrial robot Give Scope to the Advantage in social services and industrial production. However, due to long-term mechanical wear and structural deformation, the absolute positioning accuracy is low, which greatly hinders the development of manufacturing industry. Calibrating the kinematic parameters of the robot is an effective way to address it. However, the main measuring equipment such as laser trackers and coordinate measuring machines are expensive and need special personnel to operate. Additionally, in the measurement process, due to the influence of many environmental factors, measurement noises are generated, which will affect the calibration accuracy of the robot. Basing on these, we have done the following work: a) developing a robot calibration method based on plane constraint to simplify measurement steps; b) employing Square-root Culture Kalman Filter (SCKF) algorithm for reducing the influence of measurement noises; c) proposing a novel algorithm for identifying kinematic parameters based on SCKF algorithm and Levenberg Marquardt (LM) algorithm to achieve the high calibration accuracy; d) adopting the dial indicator as the measuring equipment for slashing costs. The enough experiments verify the effectiveness of the proposed calibration algorithm and experimental platform.

*Index Terms*—Industrial robots, Robot Calibration, Positioning Error, Square-Root Culture Filter Algorithm, Levenberg-Marquardt Algorithm.

## I. Introduction

IN THE FIELD OF CONTEMPORARY INDUSTIAL MANUFACTURING, industrial robots are widely used in electronic equipment, semiconductors, printed circuit boards, automobiles and other fields for handling and assembly, which are indispensable intelligent automatic mechanical equipment for advanced industrial production [1]-[11]. However, due to the deformation and wear of industrial robot, industrial robot's absolute positioning accuracy inevitably decreases [10]-[13], which is hard to keep up with the demand of high-accuracy industrialization [1], [13]-[19].

Kinematic calibration is an effective means to enhance the industrial robot's absolute positioning accuracy. In this work, since the kinematic parameters error accounts for 90% of all errors of the robot, the dynamic error is not considered [20]-[28]. The accurate measuring device is necessary for kinematic calibration such as laser tracker and coordinate measuring machines [29]. However, these measuring devices is expensive, and they need professional personnel to operate. Dial indicator is a high-accuracy and low-cost equipment, which is employed in various measurement environments.

The kinematic parameters identification is the process of obtaining the actual parameters of the robot kinematic model by calculating the actual measured values of the position and the theoretical position of the end effector. [25], [27]. Liu *et al.* [22] design a modified Simulated Annealing algorithm (MSAA), which devise a suitable cooling schedule that can accelerate the convergence rate. Alıcı *et al.* [30] employs the Particle Swarm Optimization (PSO) algorithm to identify the kinematic parameters of robot, which heightened the robot calibration accuracy. Levenberg Marquardt (LM) algorithm [29] is a nonlinear identification method combining gradient descent method and Newton method with fast calculation speed and strong convergence, which can enhance the performance of robot with effect [31]-[34]. Moreover, Square-Root Culture Kalman Filter (SCKF) algorithm [35] is an improved algorithm of Cubature Kalman Filter algorithm (CKF), which can effectively address the problem of filtering result divergence caused by negative determination of error covariance matrix ensuring the numerical stability of filtering algorithm and heightening tracking accuracy and reliability.

In this work, SCKF is firstly employed to deal with the data noise, and then the kinematic parameters are further optimized by LM algorithm. Furthermore, adopting the dial indicator as the measuring equipment, a robot error compensation method based on plane constraint is proposed to achieve high-accuracy robot calibration and low cost.

The contributions of this work are as follows: a) Developing a robot calibration method based on plane constraint to heighten robot calibration efficiency; b) Adopting SCKF algorithm for reducing the influence of measurement noises; c) Proposing a novel method for identifying kinematic parameters based on SCKF and LM algorithm. d) Adopting the dial indicator as the measuring equipment for slashing costs.

Section II states robot kinematic, kinematic error model and kinematic error identification model. Section III proposes a novel method for identifying kinematic parameters. Section IV reports the experimental results and the conclusions are illustrated in Section V.

## II. Preliminaries

For the research of robot kinematics, D-H model including $\theta_i$ (joint Angle), $a_i$ (link length), $d_i$ (link offset), $\alpha_i$ (link twist angle) is employed to describe the pose of robot end-effector generally. Table I shows the nominal D-H parameters of industrial robots.

TABLE I. ABB IRB120 Industrial robot D-H parameters.

| Joint $i$ | $\alpha_i/°$ | $a_i/mm$ | $d_i/mm$ | $\theta_i/°$ |
|---|---|---|---|---|
| 1 | -90 | 249.85003 | 653.5 | 0 |
| 2 | 0 | 900.32123 | 0 | -90 |
| 3 | 90 | -0.20614449 | 0 | 180 |
| 4 | -90 | 0 | 1030.37534 | 0 |
| 5 | 90 | 0 | 0 | 90 |
| 6 | 0 | 0 | 200.6 | 0 |

The homogeneous coordinate transformation relationship between adjacent links of robot is as follow:

$$T_i = Rot_Z(\theta_i) \times Trans_Z(d_i) \times Trans_X(a_i) \times Rot_X(\alpha_i) = \begin{bmatrix} c\theta_i & -s\theta_i c\alpha_i & s\theta_i s\alpha_i & a_i c\theta_i \\ s\theta_i & c\theta_i c\alpha_i & -c\theta_i s\alpha_i & a_i s\theta_i \\ 0 & s\alpha_i & c\alpha_i & d_i \\ 0 & 0 & 0 & 1 \end{bmatrix}, \quad (1)$$

where $c\theta_i$ and $s\theta_i$ represent $\cos\theta$, $\sin\theta$, respectively.

According to the robot forward kinematics, the homogeneous transformation relationship between the end-effector coordinate system and the robot base coordinate system is obtained as:

$$T = T_1 T_2 T_3 T_4 T_5 T_6. \quad (2)$$

*A. Kinematic Error model*

According to the differential principle, we achieve the differential transformation relationship between adjacent links as:

$$dT_i = \frac{\partial T_i}{\partial \alpha_i}\Delta\alpha_i + \frac{\partial T_i}{\partial a_i}\Delta a_i + \frac{\partial T_i}{\partial d_i}\Delta d_i + \frac{\partial T_i}{\partial \theta_i}\Delta\theta_i = T_i \delta T_i. \quad (3)$$

Considering the kinematic parameter errors of each link of the robot, the kinematic error model is established as:

$$T + dT = \prod_{i=1}^{6}(T_i + dT_i). \quad (4)$$

Then we can get the kinematic errors as:

$$\Delta P = \begin{bmatrix} J_1 & J_2 & J_3 & J_4 \end{bmatrix} \begin{bmatrix} \Delta\alpha \\ \Delta a \\ \Delta d \\ \Delta\theta \end{bmatrix} = J\Delta x, \quad (5)$$

where $\Delta\alpha$, $\Delta a$, $\Delta d$, $\Delta\theta$ are the vector of D-H parameter deviations, each of which compose of six parameters, $J$ is the Jacobian matrix related to D-H parameters. $\Delta X$ is the error of kinematic parameters, and $\Delta P$ is the position error of the end-effector.

*B. Kinematic Error Identification model*

In this work, adopting the dial indicator as the measuring equipment, we propose an error compensation method based on plane constraint. The modeling principle is that to establish a linear relationship. The modeling principle is that the measurement points meet the properties of the same plane in theory, so the robot error identification model with plane constraint is established. The schematic diagram of modeling principle based on plane constraint is demonstrated in Fig. 1.

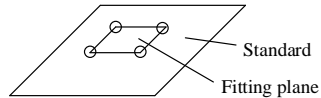

Fig. 1. The schematic diagram of modeling principle based on plane constraint.

Due to the low positioning accuracy of the robot, the measured nominal position points of the robot end-effector are not in the standard plane. Therefore, errors can occur between the plane fitted by any three points of the nominal position points obtained by the robot control software and the actual plane. The mathematical expression of the plane equation is can be described as:

$$(P-\gamma)\beta^T = 0, \quad (6)$$

where $\gamma$ is a point on the plane, $\beta$ is the plane normal vector, $P=[X, Y, Z]$ is a actual position, which can be expressed as:

$$\begin{aligned} X &= P'_x + J_x \Delta x, \\ Y &= P'_y + J_y \Delta x, \\ Z &= P'_z + J_z \Delta x, \end{aligned} \quad (7)$$

where $P=[P_x, P_y, P_z]$ is the nominal position, and $J=[J_x, J_y, J_z]$ is the identification Jacobian matrix.

Additionally, due to the error between the fitting plane and the standard plane, the plane normal vector $\beta$ exists the error $\Delta\beta$. Then, we put the coordinates of actual position of the robot end-effector in to equation (6) to obtain the equation as:

$$J\beta^T \Delta x + P' \Delta \beta^T - \gamma \Delta \beta^T = -(P' - \gamma)\beta^T \Leftrightarrow \begin{bmatrix} J\beta^T & P' & 1 \end{bmatrix} \begin{bmatrix} \Delta x \\ \Delta \beta^T \\ -\gamma \Delta \beta^T \end{bmatrix} = -(P' - \gamma)\beta^T, \quad (8)$$

where matrix $J\beta$ and $-(P'-\gamma_0)\beta^T$ can be obtained from the nominal position points of the robot end-effector, $\Delta x = [\Delta\alpha, \Delta a, \Delta d, \Delta\theta]$ is the D-H parameters error matrix.

With the above analysis, we can define the objective function as:

$$f = \frac{1}{2N} \sum_i^N \left\| (P' - \gamma)\beta^T \right\|_2^2, \quad (9)$$

where $N$ is the number of the collected samples.

### III. KINEMATIC PARAMETER IDENTIFICATION

*C. Square Root Culture Kalman Filter (SCKF)*

With kinematic error model and identification model, we can obtain the state equation of SCKF algorithm as:

$$X_{i,k-1|k-1} = S_{k-1|k-1}\zeta_i + x'_{k-1|k-1}, \quad (10)$$

$$X^*_{i,k|k-1} = X_{i,k-1|k-1}, \quad (11)$$

$$x'_{k|k-1} = \frac{1}{m}\sum_{i=1}^m X^*_{i,k|k-1}, \quad (12)$$

$$\chi^*_{k|k-1} = \frac{1}{\sqrt{m}}\left[ X^*_{1,k|k-1} - x'_{k|k-1}, X^*_{2,k|k-1} - x'_{k|k-1}, \ldots X^*_{m,k|k-1} - x'_{k|k-1} \right], \quad (13)$$

$$S_{k|k-1} = Tria\left( \left[ \chi^*_{k|k-1}, S_Q \right] \right), \quad (14)$$

where $m$ is the number of volume points, $\zeta=(m/2)^{1/2}[I]_i$ is the volume points set, and $[I]_i$ represents the unit matrix. $X_{i,k}$ is the state volume point, $X^*_k$ the volume point after transition. $x'_k$ is the state prediction value, and $\chi^*_k$ represent the square root of the state covariance matrix. With equation (10), when $k=1$, the initial covariance matrix $P_0$ is decomposed by Cholesky decomposition to obtain $S_{(0)k}=chol(P_{(0)k})$; when $k \geq 2$, the error matrix is $S_{k-1|k-1}=S_{k-1}$, which is decomposed by QR decomposition. With equation (14), $Trial(\cdot)$ means QR decomposition of the transpose of the matrix of order $A_1 \times A_2$ ($A_1 < A_2$). The first part $M \times M$ in the R matrix is taken to obtain $A_1 \times A_1$ matrix. $S_Q$ is the root mean square matrix of the state noise covariance matrix $Q_K$.

The measurement update equations are as follow:

$$X_{i,k+1|k} = S_{k|k-1}\xi_i + x'_{k|k-1}, \quad (15)$$

$$Z_{i,k|k-1} = X_{i,k|k-1}, \quad (16)$$

$$z'_{k,|k-1} = \frac{1}{m}\sum_i^m Z_{i,k|k-1}, \quad (17)$$

$$\zeta_{k|k-1} = \frac{1}{\sqrt{m}}\left[ Z_{1,k-1|k-1} - z'_{k,|k-1}, Z_{2,k-1|k-1} - z'_{k,|k-1}, \ldots, Z_{m,k-1|k-1} - z'_{k,|k-1} \right], \quad (18)$$

$$S_{ZZ,k|k-1} = Tria\left( \left[ \zeta_{k|k-1}, S_R \right] \right), \quad (19)$$

$$\Gamma_{k|k-1} = \frac{1}{\sqrt{m}}\left[ X_{1,k-1|k-1} - x'_{k|k-1}, X_{2,k-1|k-1} - x'_{k|k-1}, \ldots, X_{m,k-1|k-1} - x'_{k|k-1} \right], \quad (20)$$

$$P_{XZ,k|k-1} = \Gamma_{k|k-1}\zeta^T_{k|k-1}, \quad (21)$$

$$K_k = P_{XZ,k|k-1}(S_{ZZ,k|k-1}S^T_{ZZ,k|k-1})^{-1}, \quad (22)$$

$$x'_{k|k} = x'_{k|k-1} + K_k(Z_{k|k} - \hat{Z}_{k|k-1}), \quad (23)$$

$$S_{k|k} = Tria\left( \left[ \Gamma_{k|k-1} - K_k\zeta_{k|k-k}, K_kS_R \right] \right), \quad (24)$$

where $z'_k$ is the measurement prediction value, $S_{ZZ,k}$ is the prediction covariance matrix, $S_R$ is the root mean square matrix of the measurement noise covariance matrix $R_k$, $P_{XZ,k}$ is the covariance between the value of the state prediction and the measurement prediction value, and $K_k$ is the Kalman gain.

*D. Levenberg Marquardt algorithm (LM)*

In this paper, we discuss the update steps of LM algorithm for robot accuracy calibration based on plane constraints. Firstly, According to the principle of LM algorithm and equation (9), we can define the plane equation as $-(P'-\gamma_{t,0})\beta_{t,0}^T = \varphi(w^r_t, \gamma_{t,0}, \beta_{t,0})$. Then the objective function can be described as:

$$\arg\min f = \frac{1}{2N}\sum_{i}^{N}\left\|\varphi_i\left(w_t^r,\gamma_{t,0},\beta_{t,0}\right)+\Delta\varphi_i\left(w_t^r,\gamma_{t,0},\beta_{t,0}\right)\right\|_2^2 + \frac{\lambda}{2}\left[\left(w_{t+1}^r - w_t^r\right)^2 + \left(\gamma_{t+1,0} - x_{t,0}\right)^2 + \left(\beta_{t+1,0} - \beta_{t,0}\right)^2\right], \quad (25)$$

With the equation (25), $\Delta\varphi(w^r_t, \gamma_{t,0}, \beta_{t,0})$ is as follow:

$$\Delta\varphi_i\left(w_t^r,\gamma_{t,0},\beta_{t,0}\right) = \frac{\partial\varphi_i}{\partial w_t^r}\left(w_{t+1}^r - w_t^r\right) + \frac{\partial\varphi_i}{\partial\gamma_{t,0}}\left(\gamma_{t+1,0} - \gamma_{t,0}\right) + \frac{\partial\varphi_i}{\partial\beta_{t,0}}\left(\beta_{t+1,0} - \beta_{t,0}\right), \quad (26)$$

where $w$ is the kinematic parameters. $\gamma_0$ and $\beta_0$ represents the initial value of the point on the plane and the plane normal vector, respectively. Moreover, $w^r_t$, $\gamma_{t,0}$, $\beta_{t,0}$ are the prior estimates, and $w^r_{t+1}$, $\gamma_{t+1,0}$, $\beta_{t+1,0}$ are the post estimates. the *r-th* column vectors are written as $w^r$. $\lambda$ is a learning coefficient for the substantial term. $I$ is an unit matrix.

With the equation (10), we can obtain the update steps as:

$$w_{t+1}^r = w_t^r + \left[\frac{1}{n}\sum_{i=1}^{n}\left(\frac{\partial\varphi_i}{\partial w_t^r}\right)^T \frac{\partial\varphi_i}{\partial w_t^r} + \lambda I\right]^{-1} \frac{1}{n}\sum_{i=1}^{n}\left(\varphi_i \frac{\partial\varphi_i}{\partial w_t^r}\right), \quad (27)$$

$$\gamma_{t+1,0} = \gamma_{t,0} + \left[\frac{1}{n}\sum_{i=1}^{N}\left(\frac{\partial\varphi_i}{\partial\gamma_{t,0}}\right)^T \frac{\partial\varphi_i}{\partial\gamma_{t,0}} + \lambda I\right]^{-1} \frac{1}{n}\sum_{i=1}^{N}\left(\varphi_i \frac{\partial\varphi_i}{\partial\gamma_{t,0}}\right), \quad (28)$$

$$\beta_{t+1,0} = \beta_{t,0} + \left[\frac{1}{n}\sum_{i=1}^{N}\left(\frac{\partial\varphi_i}{\partial\beta_{t,0}}\right)^T \frac{\partial\varphi_i}{\partial\beta_{t,0}} + \lambda I\right]^{-1} \frac{1}{n}\sum_{i=1}^{N}\left(\varphi_i \frac{\partial\varphi_i}{\partial\beta_{t,0}}\right). \quad (29)$$

## IV. XPERIMENTAL RESULTS

### E. Experimental Setup

The effectiveness of the presented method is verified by experiments in this section. Fig. 2 illustrates the experimental platform, which mainly contains a HSR-JR680 industrial robot, a dial indicator, a high accuracy gauge block and a PC.

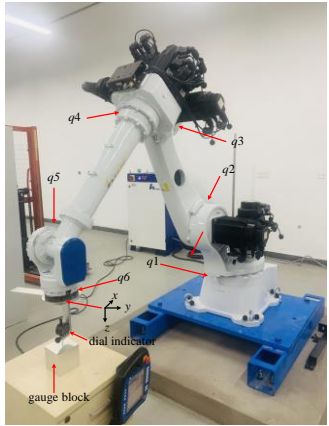

Fig. 2. The experimental platform.

### F. Evaluation Metrics

We employ maximum error (MAX) [29], [36]-[40], standard deviation (STD) and root mean squared error (RMSE) are adopted as the evaluation metrics in the industrial robot calibration experiments:

$$MAX = \max\left(\sqrt{\left(\left(P'-\gamma\right)\beta^T\right)^2}\right), \quad (29)$$

$$STD = \frac{1}{N}\sum_{i=1}^{N}\left(\sqrt{\left(\left(P'-\gamma\right)\beta^T\right)^2}\right), \quad (30)$$

$$RMSE = \sum_{i=1}^{N}\left(\sqrt{\frac{1}{N}\left(\left(P'-\gamma\right)\beta^T\right)^2}\right). \quad (31)$$

### G. Experimental Process

In this work, we collected 120 samples on the plane of the gauge block, which will cover the whole plane of the gauge block as much as possible. In this process, the dial indicator reading corresponding to each sample is recorded. After completing the sample collection, dial indicator readings are added to the link offset of the sixth axis of the robot. Then we employ the proposed calibration method to identify the kinematic parameters error, and finally the error compensation is done. All calibration algorithms are shown in Table II.

TABLE II. Compared algorithms.

| Model | Description |
|---|---|
| M1 | Simulated Annealing algorithm (SA) [28], basing probability, originates from the principle of solid annealing, which can search for the global solution. |
| M2 | Particle Swarm Optimization algorithm (PSO) [30] is a biological heuristic method with high accuracy and fast convergence speed. |
| M3 | Extended Kalman Filter algorithm (EKF) [1] is an extended form of standard Kalman Filter in nonlinear situations, which is an efficient recursive filter. |
| M4 | Differential Evolution Algorithm (DE)[20] is a heuristic random search algorithm based on population differences, which has strong robustness. |
| M5 | Square-Root Culture Kalman Filter (SCKF) algorithm [35] is an improved algorithm of the CKF algorithm, which can effectively address the problem of filtering result divergence caused by negative determination of error covariance matrix. |
| M6 | Levenberg Marquardt algorithm (LM) [20], [29] is the most widely used nonlinear least square algorithm, which has the advantages of both gradient method and Newton method. |
| M7 | The proposed SCKF+LM algorithm, which is adopted to enhance the calibration accuracy of industrial robot. |

*H. Experimental Performance of Compared systems*

We verify the performance of SCKF+LM algorithm in dealing with industrial robot calibration by carefully comparing with four advanced algorithms. The calibration accuracy of all compared algorithms are demonstrated in Table III. Furthermore, Table IV shows the calibration accuracy of all algorithms, and the time costs are shown in Table V. Fig. 3, Fig. 4 and Fig. 5 illustrate the calibration accuracy, training curves and the positional errors of these calibration algorithms after calibration, respectively. We can draw the following conclusions:

a) **M7 has the highest calibration accuracy.** As shown in Table III and Fig. 3, the RMSE, STD and max of M7 are 0.49mm, 0.40mm and 1.14mm respectively, which are 15.51%, 18.36% and 9.92% higher than those of M6 respectively. Therefore, this algorithm is an effective method for robot accuracy calibration, which can greatly enhance the accuracy of robot calibration.

TABLE III. Calibration accuracy of all compared algorithms.

| Model | RMSE(mm) | STD(mm) | MAX(mm) |
|---|---|---|---|
| Before Calibration | 2.28 | 2.15 | 4.09 |
| M1 | 0.72 | 0.65 | 1.83 |
| M2 | 0.65 | 0.56 | 1.71 |
| M3 | 0.97 | 0.83 | 2.23 |
| M4 | 0.60 | 0.55 | 1.57 |
| M5 | 0.69 | 0.54 | 1.72 |
| M6 | 0.58 | 0.49 | 1.31 |
| M7 | 0.49 | 0.40 | 1.18 |

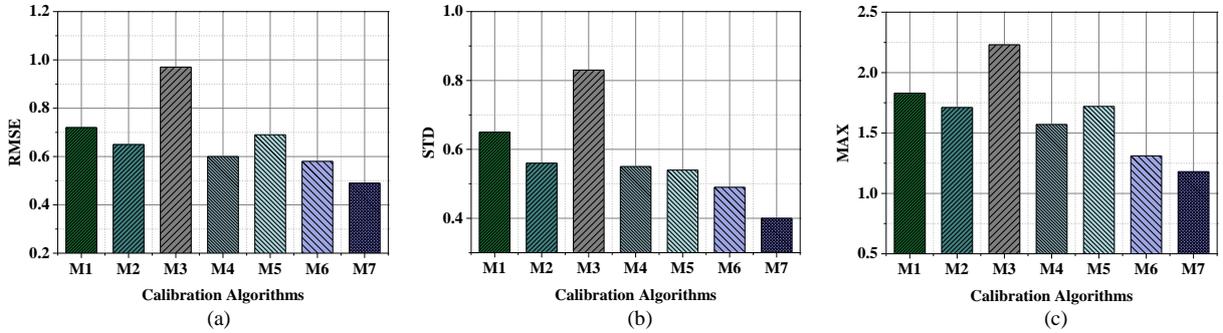

Fig. 3. The calibration results of M1-M7. (a) The RMSE of compared algorithms. (b)The STD of compared algorithms. (c)The MAX of compared algorithms.

b) **M5's convergence speed is competitive.** As depicted in Fig. 4(a) and (b), among M1-6, M3 has the fastest convergence speed (only 20 iterations). However, from Fig. 3, M3 the worst robot calibration accuracy. In comparison, M5 has better robot calibration accuracy, and it only needs 35 iterations to converge. Thus, we employ SCKF algorithm to preprocess the data noise, rather than EKF algorithm.

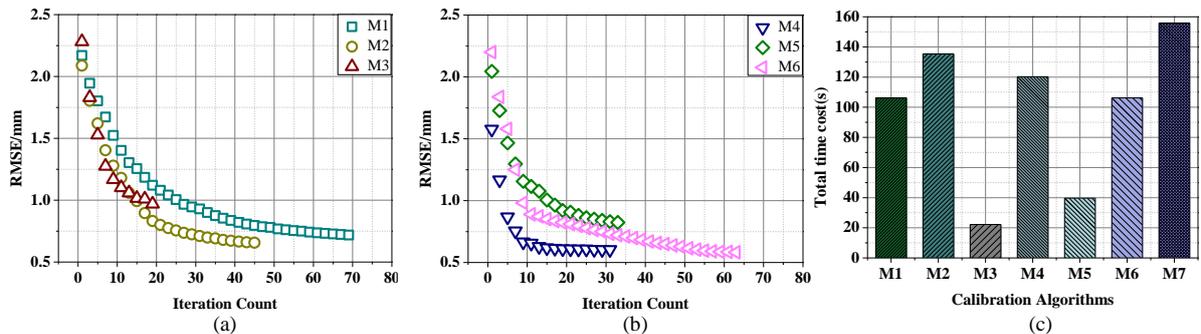

Fig. 4. Training curves and total time cost of algorithms M1-M7. (a) Training curves of M1, M2, M3. (b) Training curves of M4, M5, M6. (c) Time of M1-M7.

c) **Although M7 reduces the computational efficiency to some extent, it greatly improves the calibration accuracy.** From Fig.4 (c) and Table IV, M3 has the highest computational performance among compared algorithms, which converges about only 22s. Moreover, comparing with M1-6, M7 has lower computational performance, which shows that the proposed SCKF+LM algorithm will damage the computational performance. However, due to effective improvement for robot calibration accuracy, the algorithm is reasonable.

TABLE IV. Time cost of all compared algorithms.

| Model | Total Time Cost(s) (RMSE) | Average Iteration Count (RMSE) |
|---|---|---|
| M1 | 106.23 | 69 |
| M2 | 135.36 | 45 |
| M3 | 22.15 | 20 |
| M4 | 120.21 | 31 |
| M5 | 39.66 | 35 |
| M6 | 106.24 | 63 |
| M7 | 155.90 | 45 |

d) **M7 significantly outperforms its peers in limiting position error for robot calibration.** Table V illustrates the kinematic parameters of HSR-JR680 industrial robot after M7 calibration. In this work, we selected 35 samples of 120 samples collected to evaluate the position error of the robot corresponding to each algorithm. From Fig. 5, M1-M6 can heighten the calibration accuracy of the robot, and the position error of M7 is obviously higher than that of M1-6. Based on the above analysis, the effectiveness of the proposed method is verified.

TABLE V. Industrial robot kinematic parameters after calibration.

| Joint $i$ | $α_i/°$ | $a_i/mm$ | $d_i/mm$ | $θ_i/°$ |
|---|---|---|---|---|
| 1 | -89.9960 | 249.7252 | 654.2394 | 0 |
| 2 | -0.0232 | 893.9318 | -0.1907 | -90 |
| 3 | 90.0073 | -210.7932 | -0.1926 | 180 |
| 4 | -90.0214 | 4.6549 | 1034.0154 | 0 |
| 5 | 90.0094 | -0.3347 | 0.1219 | 90 |
| 6 | 0 | -0.0534 | 199.8700 | 0 |

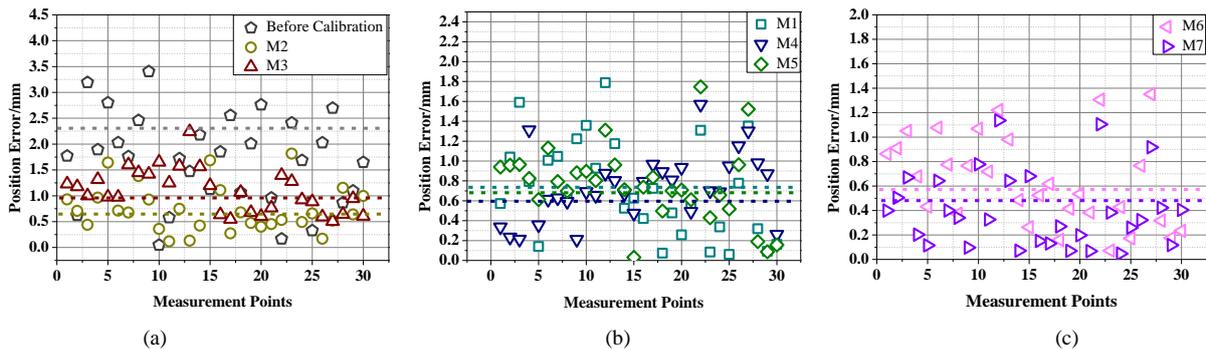

(a)      (b)      (c)

Fig. 5. The robot positioning errors after calibration through models M1-M7. (a) Position errors of before calibration, M2, M3. (b) Position errors of M1, M4, M6. (c) Position errors of M6, M7. It shows that M7 has the best calibration accuracy.

## V. CONCLUSIONS

A novel robot calibration algorithm based on SCKF and LM algorithm is proposed for enhancing the robot calibration accuracy and reducing the noise impact. Moreover, the dial indicator as the measuring equipment, we develop a plane constraint method to heighten robot calibration efficiency. Additionally, employed dial indicator instead of laser tracker or coordinate measuring machines as the measuring equipment can greatly save cost for robot calibration. The enough experiments verify the effectiveness of the proposed experimental platform and calibration algorithm.

For the future work, we will investigate the other measuring equipment such as laser radar and depth camera to achieve the online calibration of industrial robot.